\documentclass[sigconf]{acmart}
\AtBeginDocument{%
  }

\copyrightyear{2026}
\acmYear{2026}
\setcopyright{cc}
\setcctype{by} 

\acmConference[WWW '26]{Proceedings of the ACM Web Conference 2026}{April 13--17, 2026}{Dubai, United Arab Emirates}
\acmBooktitle{Proceedings of the ACM Web Conference 2026 (WWW '26), April 13--17, 2026, Dubai, United Arab Emirates}
\acmISBN{979-8-4007-2307-0/2026/04}
\acmDOI{10.1145/XXXXXX.XXXXXX}

\usepackage{hyperref}
\usepackage{url}
\usepackage{enumitem}
\usepackage{graphicx}
\usepackage{booktabs,makecell,adjustbox}
\usepackage{multirow} 
\usepackage{adjustbox}
\usepackage{amsmath}
\usepackage{tcolorbox}
\tcbuselibrary{breakable}
\usepackage{array}
\newcolumntype{C}[1]{>{\centering\arraybackslash}m{#1}}

\usepackage{tcolorbox}
\tcbuselibrary{listings,breakable}
\usepackage{listings}
\definecolor{darkred}{rgb}{0.6,0.0,0.0}
\definecolor{darkgreen}{rgb}{0,0.50,0}
\definecolor{lightblue}{rgb}{0.0,0.42,0.91}
\definecolor{orange}{rgb}{0.99,0.48,0.13}
\definecolor{grass}{rgb}{0.18,0.80,0.18}
\definecolor{pink}{rgb}{0.97,0.15,0.45}
\definecolor{codegreen}{rgb}{0,0.6,0}
\definecolor{codegray}{rgb}{0.5,0.5,0.5}
\definecolor{codepurple}{rgb}{0.58,0,0.82}
\definecolor{backcolour}{rgb}{0.95,0.95,0.92}

\newcommand{\best}[1]{{\color{darkgreen}\textbf{#1}}}  
\newcommand{\second}[1]{\underline{#1}}

\begin{document}

\title{Matrix as Plan: Structured Logical Reasoning with Feedback-Driven Replanning}

\settopmatter{authorsperrow=3}

\author{Ke Chen}
\authornote{Equal contribution.}
\email{kechen@mail.bnu.edu.cn}
\affiliation{%
   \department{Faculty of Arts and Sciences}
   \institution{Beijing Normal University}
  \city{Zhuhai}
  \state{Guangdong}
  \country{China}
}

\author{Jiandian Zeng}
\authornotemark[1] 
\email{jiandian@bnu.edu.cn}
\affiliation{%
   \department{Institute of Artificial Intelligence and Future Networks}
   \institution{Beijing Normal University}
  \city{Zhuhai}
  \state{Guangdong}
  \country{China}
}

\author{Zihao Peng}
\email{pzh_cs@mail.bnu.edu.cn}
\affiliation{%
   \department{Institute of Artificial Intelligence and Future Networks}
   \institution{Beijing Normal University}
  \city{Zhuhai}
  \state{Guangdong}
  \country{China}
}

\author{Guo Li}
\email{liguo@mail.bnu.edu.cn}
\affiliation{%
   \department{Institute of Artificial Intelligence and Future Networks}
   \institution{Beijing Normal University}
  \city{Zhuhai}
  \state{Guangdong}
  \country{China}
}

\author{Guangxue Zhang}
\email{guangxue_zhang@bnu.edu.cn}
\affiliation{%
   \department{Institute of Artificial Intelligence and Future Networks}
  \institution{Beijing Normal University}
  \city{Zhuhai}
  \state{Guangdong}
  \country{China}
}

\author{Tian Wang}
\authornote{Corresponding author.}
\email{tianwang@bnu.edu.cn}
\affiliation{%
   \department{Engineering Research Center of Cloud-Edge Intelligent Collaboration on Big Data, Ministry of Education}
   \institution{Beijing Normal University}
  \city{Zhuhai}
  \state{Guangdong}
  \country{China}
}

\begin{abstract}

As knowledge and semantics on the web grow increasingly complex, enhancing Large Language Models (LLMs)' comprehension and reasoning capabilities has become particularly important. Chain-of-Thought (CoT) prompting has been shown to enhance the reasoning capabilities of LLMs. However, it still falls short on logical reasoning tasks that rely on symbolic expressions and strict deductive rules. Neuro-symbolic methods address this gap by enforcing formal correctness through external solvers. Yet these solvers are highly format-sensitive, and small instabilities in model outputs can lead to frequent processing failures. The LLM-driven approaches avoid parsing brittleness, but they lack structured representations and process-level error-correction mechanisms. To further enhance the logical reasoning capabilities of LLMs, we propose MatrixCoT, a structured CoT framework with a matrix-based plan. Specifically, we normalize and type natural language expressions and attach explicit citation fields, and introduce a matrix-based planning method to preserve global relations among steps. The plan thus becomes a verifiable artifact and execution becomes more stable. For verification, we also add a feedback-driven replanning mechanism. Under semantic-equivalence constraints, it identifies omissions and defects, rewrites and compresses the dependency matrix, and produces a more trustworthy final answer. Experiments on five logical-reasoning benchmarks and five LLMs show that, without relying on external solvers, MatrixCoT enhances both the robustness and interpretability of LLMs when tackling complex symbolic reasoning tasks, while maintaining competitive performance.

\end{abstract}

\ccsdesc[500]{Computing methodologies}
\ccsdesc[500]{Computing methodologies~Artificial intelligence}
\ccsdesc[500]{Computing methodologies~Natural language processing}

\keywords{Logical Reasoning, Large Language Models, Neurosymbolic Approaches, Semantic Decomposition}

\renewcommand{\shortauthors}{Ke Chen et al.}
\maketitle

\section{Introduction}

The Web's explosive growth has created a vast, heterogeneous substrate of knowledge. Extracting task-relevant content and reliable semantic knowledge from these repositories is essential~\cite{penedo2024fineweb, web1, web2}. This necessity has motivated the integration of Large Language Models (LLMs) into web-based applications, including browsing-based question answering, cross-site web agents, and LLM-assisted information extraction and knowledge-graph construction~\cite{nakano2021webgpt, yoran2024assistantbench, xu2024large, chen2023autokg}. Despite their strong capabilities in natural language generation, text classification, and machine translation~\cite{achiam2023gpt4, liu2024deepseek_v3, comanici2025gemini}, current LLMs remain unreliable on complex logical reasoning~\cite{shojaee2025illusion, roh2025chain}. The gap is most evident in higher-order tasks such as mathematical problem solving and intricate logical inference. Consequently, to ensure trustworthy behavior in web applications, developing robust reasoning is essential for advancing LLMs within the Web's semantics and knowledge ecosystem.

\begin{figure*}[t]
  \centering
  \includegraphics[width=\linewidth, keepaspectratio]{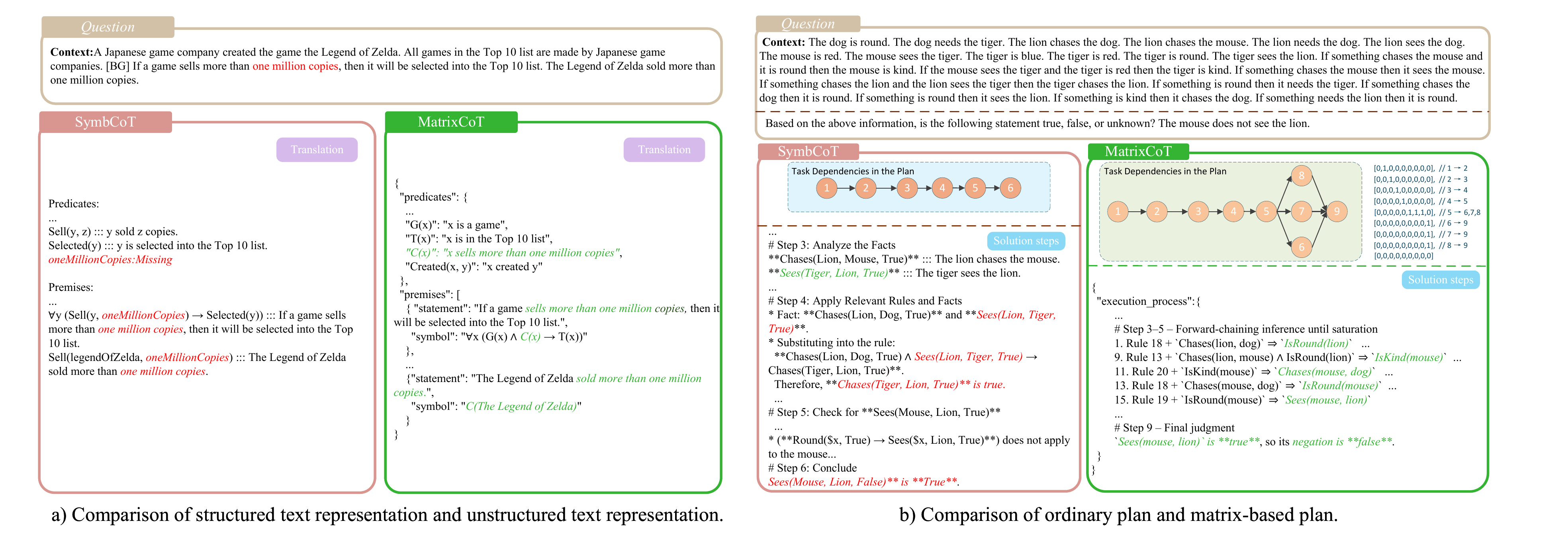}
  \caption{Illustration of Logical Reasoning with SymbCoT and the Proposed MatrixCoT. a) Structured vs. unstructured text representation. b) Ordinary plan vs. matrix-based plan.}

  \label{fig1}
  
\end{figure*}

Motivated by this need, a growing body of work now examines how LLMs perform on such reasoning tasks. Current LLMs are pretrained primarily on human-written natural text, where the corpora lack high-quality examples of logical reasoning, deduction, and proof~\cite{frieder2024data, wang2024mathpile, morishita2024enhancing}. Moreover, prevailing training objectives, such as next token prediction or masked language modeling, bias models toward learning syntax, semantics, and commonsense knowledge~\cite{luo2023towards}. As a result, even state-of-the-art LLMs still struggle to achieve satisfactory performance on complex logical reasoning tasks. To improve the reasoning capacity, researchers have proposed Chain-of-Thought (CoT) prompting, which encourages step-by-step decomposition of complex problems~\cite{wei2022chain}. Variants such as Tree of Thoughts (ToT)~\cite{yao2023tree} and Graph of Thoughts (GoT)~\cite{besta2024graph} further extend this idea. While these methods enhance reasoning to some degree, they remain limited by the abstraction and vagueness of natural language. Genuine logical reasoning, however, requires symbolic representations and formal deduction rules, which are largely absent from textual corpora. The tension is especially pronounced in settings that demand rigorous logical representations and formal derivations.

To address the above limitations, researchers have explored several neuro-symbolic approaches. One prominent line of work emphasizes a hybrid architecture that integrates an LLM with a symbolic reasoning backend~\cite{gao2023pal, ryu2024divide}: 1) The LLM first translates a natural language statement of the reasoning problem into a set of first-order logic formulas;
and 2) a symbolic solver uses this formal representation to automatically plan the reasoning steps, execute the derivations, and produce the final answer.
Within this pipeline, the symbolic solver plays a pivotal role in both planning and execution. It imposes constraints and performs verification to mitigate reasoning errors that arise from the inherent abstraction and vagueness of natural language formulations~\cite{ye2023satlm, pan2023logic, kirtania2024logic, olausson2023linc}.

However, external reasoners require extremely strict formatting on their inputs. Even slight deviations introduced by an LLM during generation can cause parsing to fail and the entire reasoning pipeline to halt. To fill the gap, LLM-driven reasoning methods~\cite{xu2024faithful, xu2024aristotle, liu2025logic} are designed to be robust to syntactic noise and structural perturbations. A representative example is SymbCoT~\cite{xu2024faithful}: rather than relying on any symbolic solver, the LLM is adopted in a systematic four-stage pipeline: translation, planning, solving, and verification, suggesting that LLM–driven reasoning frameworks offer strong generative flexibility. However, in practice, they exhibit limitations that may degrade reasoning performance:

\begin{itemize}[nosep, leftmargin=*]
\item[1)] \textbf{Reasoning by LLMs on prose-style text intermixes predicates, entities, and premises within free-form narration.} It maintains only loose alignment through inline annotations (e.g., \texttt{:::}). In the absence of a unified namespace and field-level constraints, types and variable scope can not be statically checked, and cross-sentential references are prone to drift. As shown in Fig.~\ref{fig1}(a), in an unstructured rendering without a unified namespace, an expression such as \( \mathrm{Sell}(y,\mathrm{oneMillionCopies}) \) inserts oneMillionCopies as a bare constant in the second argument. This token is neither defined in the predicate inventory nor annotated with a type or a unit. Consequently, the rule \( \forall y\,\big(\mathrm{Sell}(y,\mathrm{oneMillion}\)\-\(\mathrm{Copies})
\rightarrow \mathrm{Selected}(y)\big) \) can only be treated passively as a ``special-case match,'' rather than expressing the computable semantics \( \text{sales} \ge 10^{6} \). Signature constraints and static checks therefore cannot be enforced. The system cannot promptly or reliably detect and constrain such semantic and typing omissions, nor can it proactively require completion. Subsequent reasoning then runs on incomplete dependencies, making unstable or erroneous outcomes likely. By contrast, structured symbolic translation, under a unified namespace, uses keys and indices for precise addressing. It can perform static checks and consistency validation for type consistency, variable scope, and cross-sentential references. With equivalent informational content, it thus yields higher decidability and overall accuracy.

\item[2)] \textbf{Linearized planning struggles to explicitly encode and schedule inter-step dependencies.} It is sensitive to wording and sampling perturbations, which often break reasoning chains and cause evidence to be missed. Take Fig. ~\ref{fig1}(b) as an example. The correct shortest reasoning chain is: Chases(lion, dog, True) → Round(lion, True); then, combined with Chases(lion, mouse, True), derive Kind(mouse, True); next, Kind(mouse, True) → Chases(mouse, dog, True) → Round(mouse, True) → Sees(mouse, lion, True). Therefore, not(Sees(mouse, lion, True)) is False. However, the executed linear plan exhibits “local inference not closed” and “semantic drift.”  At Step 4, it applies the rules only once. It then proceeds to the derivability test in Step 5 without forward-chaining to a fixed point. As a result, the knowledge base lacks key facts that should have been entailed—Kind(mouse, True), Chases(mouse, dog, True), and Round(mouse, True). Consequently, it misclassifies the query as “underivable.” It also confuses the argument direction in rule premises, conflating Sees(lion, tiger, True) with the given Sees(tiger, lion, True), which further amplifies the error.  In contrast, a matrix-based plan renders the cross-step dependency structure explicit. Reasoning paths are constrained by an adjacency matrix. Expansion of the logical chain is enforced in dependency order. It prevents skipping and omission and more systematically detects cyclic inconsistencies. Consequently, the correct answer could be obtained.
\end{itemize}

\begin{figure*}[t]
  \centering
  \includegraphics[width=\linewidth, keepaspectratio]{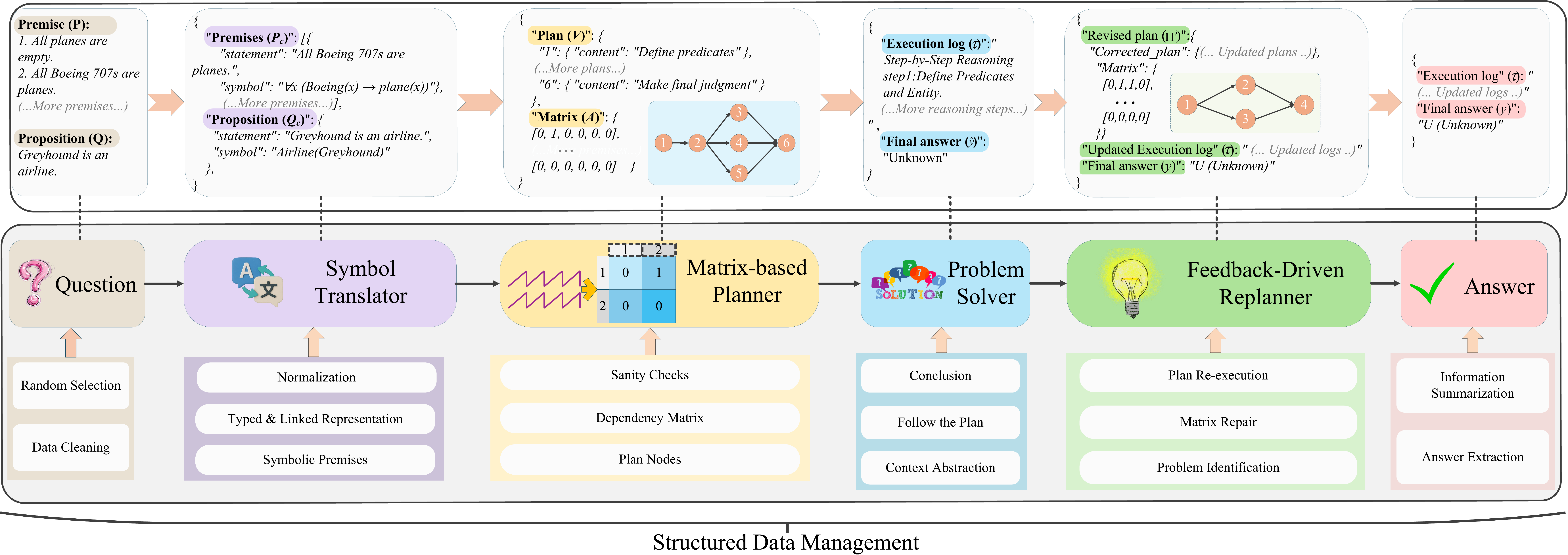}
  \caption{An overview of the MatrixCoT framework and methodology. The pipeline first translates the question into normalized, typed symbolic representations, then builds a matrix-based plan with dependencies and plan nodes. A problem solver executes the plan next, followed by a feedback-driven replanner to repair and re-execute the plan when needed. The system ultimately summarizes the information and extracts the answer.}
  \label{fig3}
  
\end{figure*}

To tackle the above issues, we propose MatrixCoT, a structured CoT approach that employs a matrix-based planning method for robust logical reasoning. The goal is to systematically enhance the reliability and controllability of LLMs on logical reasoning tasks. Unlike traditional implementations that rely on linear chains and natural language narration, MatrixCoT retains a LLM-driven paradigm while introducing structured symbolic representations, a matrix-based planning method, and a feedback-driven replanning mechanism. Specifically, we organize reasoning data into a structured symbolic form where predicates, entities, rules, and questions are encoded with explicit types and links. The reasoning plan is then represented as a dependency matrix $A$, whose rows and columns denote steps and whose entries capture prerequisite–successor relations. It turns the plan from a fragile text-based script into a globally consistent, computable structure that avoids ambiguity and preserves inter-step relations. Finally, a feedback-driven replanner examines solver traces to identify issues such as missing premises or conflicts. With minimal edits, $A$ is updated and the instance is re-solved, thereby strengthening robustness and accuracy. Our contributions can be summarized as follows:
\begin{itemize}[ itemsep=-1.5pt, topsep=0pt]
    \item \textbf{MatrixCoT.} This paper proposes MatrixCoT, a structured CoT method for logical reasoning. Compared with existing approaches, MatrixCoT organizes LLM outputs into structured representations and adopts matrix-based planning with feedback-driven replanning to guide execution more clearly, thereby achieving better robustness and interpretability on logical reasoning tasks.

    \item \textbf{Structured data management.} We introduce a method that converts natural language expressions into a normalized, typed representation with explicit referential links. This makes logical dependencies explicit, curbs semantic ambiguity, and improves both execution stability and answer quality.

    \item \textbf{Matrix-based planning method.} In this method, we represent the reasoning plan as a unified dependency matrix. This matrix preserves the global relations among steps, thereby turning the plan into a verifiable, computable artifact and further improving the consistency and controllability of the LLM reasoning process.

    \item \textbf{Feedback-driven replanning mechanism.} The system uses previous solution trajectories and candidate answers as audit signals. Under semantic equivalence constraints, omissions and defects are identified. The dependency matrix is then rewritten and compressed, thereby producing a more trustworthy final answer.
\end{itemize}

\section{Methodology}

As presented in Fig.~\ref{fig3}, the MatrixCoT framework comprises four core modules: a symbol translator, a matrix-based planner, a problem solver, and a feedback-driven replanner. In this section, the structured data management, a matrix-based planning method, and a feedback-driven replanning mechanism are detailed.
\subsection{Structured Data Management}

Recent studies show that converting the reasoning process into structured representations can alleviate the issues above~\cite{wu2025effectively}. Structured carriers reduce semantic ambiguity and explicitly maintain dependencies between pieces of information. In CoT settings, embedding intermediate reasoning steps into parseable fields significantly improves the accuracy of mathematical and symbolic reasoning. Some reports even claim that adding just a single field to carry the reasoning yields about a 60\% gain on GSM8K~\cite{leo2024bad-schemas}. Going further, the schema design itself directly affects performance~\cite{tam2024let}. Small changes in field names or ordering can cause orders-of-magnitude differences; for example, renaming ``final\_choice'' to ``answer'' can raise accuracy from 4.5\% to 95\%~\cite{leo2024bad-schemas}.

Inspired by the above research, we propose a prompt-driven structured representation that covers the full pipeline of symbolic reasoning. For illustration, in the symbol translation process, we define a parsing map from a natural language sequence \(x_{1:T}\) (a sequence of text spans of length \(T\)) to symbolic objects:
\begin{equation}
f_{\text{prompt}}:\; x_{1:T}\longrightarrow S=\langle \mathcal{V}, \mathcal{K}, \mathcal{F}, \mathcal{R}, \mathcal{Q} \rangle .
\end{equation}
Here, \(\mathcal{V}\) denotes the symbolic vocabulary; \(\mathcal{K}\) the set of constants; \(\mathcal{F}\) the set of normalized facts; \(\mathcal{R}\) the set of rules; and \(\mathcal{Q}\) the set of propositions to be adjudicated. At the symbolic layer we adopt a standard first-order logical vocabulary, for example the usual Boolean connectives, quantifiers, and equality, such that:
\begin{equation}
\{\neg,\wedge,\vee,\rightarrow,\leftrightarrow,\forall,\exists,=\}\ \subseteq\ \mathcal{L}_{\mathrm{FO}}.
\end{equation}

Finally, we serialize $S$ into a key–value form $\mathrm{enc}(S)$, defined as an alignment between Natural Language (NL) and symbolic expressions:
\begin{equation}
\mathrm{enc}(S)=\{(NL_i, Symbol_i)\mid i=1,\dots,N\}.
\end{equation}
This general mapping captures the correspondence between each natural language statement and its canonical symbolic form, facilitating downstream auditing and reasoning. Compared with free-text representations that maintain only a loosely aligned structure via inline markers (e.g., “:::”),
this representation reduces semantic ambiguity and the propagation of redundancy. It also provides stable anchors for planners and solvers. In feedback-driven settings, it further supports auditable intermediate evidence and replanning. Without sacrificing expressivity, this structured representation improves the reliability and controllability of complex symbolic reasoning.

\subsection{Matrix-based planning method}
Most contemporary reasoning-augmentation methods treat free-form text as the primary organizational unit. CoT develops the derivation through step-by-step narration. Self-Consistency stabilizes outcomes via voting across multiple chains. Least-to-Most first decomposes and then recomposes the problem. ToT seeks to orchestrate “thought units” within a tree. Representative pipeline implementations of “plan-then-solve” also fall into this family. These approaches improve performance to a degree. Yet the “plan” is still largely stitched together from textual fragments. Prerequisite–successor relations across steps are maintained mainly by implicit ordering or heuristic scores, which makes it difficult to accumulate an explicit, machine-checkable dependency structure. When backtracking, pruning, or re-execution is required, plans across samples or iterations lack strong alignment and comparability. Search-style expansions additionally introduce substantial computational overhead and instability.

Building on these observations, we introduce a matrix-based planning method. It specifies reasoning steps and their precedence constraints directly and in a computable way.  Formally, the plan is written as $\Pi=\langle S,A\rangle$: here $S=\{s_1,\ldots,s_N\}$ is a finite set of steps. Each $s_i$ denotes an inference step. $N$ is the number of steps. The matrix $A\in\{0,1\}^{N\times N}$ is a binary precedence matrix that encodes dependencies:
\begin{align}
A_{ij}=1 &\Rightarrow s_i \prec s_j,\\
A_{ij}=0 &\Rightarrow s_i \parallel s_j.
\end{align}
Here, $\prec$ means ``directly precedes.'' The symbol $\parallel$ means ``there is no direct dependency edge between the two.'' The row index $i$ marks a ``source step,'' and the column index $j$ marks a ``target step.'' Therefore, a column corresponds to the set of prerequisites required before execution, and a row corresponds to the set of successors that are unlocked after completion.

To further clarify the semantics, we define two set operations:
\begin{align}
\mathrm{pred}(s_j)=\{s_i\in S \mid A_{ij}=1\},\\
\mathrm{succ}(s_i)=\{s_j\in S \mid A_{ij}=1\}.
\end{align}
Here, $\mathrm{pred}(s_j)$ is the full set of predecessors of step $s_j$, which must be finished before executing $s_j$. $\mathrm{succ}(s_i)$ is the full set of successors of step $s_i$, which can be directly triggered once $s_i$ is done. If column $j$ is all zeros, then $\mathrm{pred}(s_j)=\varnothing$, denoting that the step has no prerequisites. If row $i$ is all zeros, then $\mathrm{succ}(s_i)=\varnothing$, signifying that the step has no successors.

Based on the above, the execution of the plan can also be formalized. At any time, select all steps $s_j$ satisfying $\mathrm{pred}(s_j)\subseteq S_{\text{done}}$ into the executable set, where $S_{\text{done}}\subseteq S$ is the set of completed steps. Then execute these steps and merge them into $S_{\text{done}}$. Repeat this process until all steps are completed or no new steps can be triggered. The execution procedure follows the structure of the matrix $A$ directly. It avoids inferring dependencies from textual descriptions and ensures clarity, robustness, and controllability of the reasoning flow.

\subsection{Feedback-Driven Replanning Mechanism}
In complex symbolic reasoning, relying solely on free-form text or a one-shot linear plan often buries key dependencies in the narrative. Premises go unused; precedence relations are obscured by the prose; evidence becomes “unreachable” along long chains. Even retries tend to fall into the same traps. We have already used a structured semantic representation to stably index predicates, entities, and rules. We also encode “which step precedes which” as an explicit dependency matrix $A$ via a matrix-based plan, so the execution side no longer has to guess dependencies. Nevertheless, an offline-generated $A$ can still be misaligned with the true difficulties of execution. Evidence produced by one step may not be consumed in time (a broken chain). Some conclusions may be consumed too early (poor timing). Parallel branches may drift out of sync, causing repeated idle cycles. A good static plan is not the same as good dynamic execution. What is missing is a channel that feeds problems exposed during execution back into the plan's structure.

To this end, we propose a feedback-driven replanning mechanism. It diagnoses and repairs historical execution traces. Let the context be $C$, the original plan $\Pi=(V,A)$ (where $V$ is the set of steps, typically ${1,\dots,N}$; each element corresponds to a planned reasoning step; $A\in{0,1}^{N\times N}$ is the dependency matrix), the execution log $\tau$, and the provisional decisions $D$. The execution trace produced by a single solve is denoted:
\begin{equation}
\tau=\langle C,\Pi,T\rangle. 
\end{equation}

Using the joint evidence from the trace and the decisions, we define a diagnostic mapping:
\begin{equation}
f_{\text{diag}}:(\tau,D)\longrightarrow \mathcal{E},
\end{equation}
where $\mathcal{E}$ is a set of failure labels. It covers categories such as ``missing prerequisites,'' ``rule misuse,'' ``premature termination,'' and ``redundancy.'' In this way, diagnosis is converted from plain text to a structured object $\bigl(V,A,\mathcal{E}\bigr)$, thus providing auditable anchors for subsequent plan editing and verification.

After issues are detected, the system performs structured edits using the dependency matrix $A$ as the unified carrier. Let $\mathcal{U}$ be the set of edit operators (e.g., $\mathrm{AddEdge}(i,j)$, $\mathrm{DelEdge}(i,j)$, $\mathrm{Merge}(p,q)$, $\mathrm{InsertGuard}(k)$, etc.). The repair is expressed as:
\begin{align}
\Pi'=(V',A')=\mathrm{Apply}\bigl(\mathcal{U},(V,A)\bigr), \\
\Pi''=\mathrm{Normalize}(\Pi'),
\end{align}
where $\mathrm{Normalize}$ removes redundancy, breaks cycles, and ensures a valid topological ordering. Re-execution depends only on ``the prior context $+$ the currently revised plan.'' The process is simple and reproducible:
\begin{equation}
\mathrm{Exec}(C,\Pi'')\longrightarrow (y,T'),
\end{equation}
where $y$ is the new answer and $T'$ is the new log. $\Pi''$ fills in the necessary prerequisites and removes spurious or transitively redundant dependencies. Normalization ensures that the main chain executes completely in topological order. With these guarantees, re-execution avoids biases such as ``premature stop,'' ``missing reasoning,'' and ``dependency errors.'' Thus, without changing the input context $C$, it yields more accurate and robust answers for complex logical reasoning.

\section{Logical Symbolic Reasoning}
\textbf{Task Definition.}
Given a set of premises $P=\{p_1,\dots,p_n\}$ (each $p_i$ is a logical sentence) and a proposition $Q$ to be decided. The possible answer is true (\(\mathbf{T}\)), false (\(\mathbf{F}\)), or unknown (\(\mathbf{U}\)). Specifically, \(\mathbf{T}\) means $Q$ can be logically derived from $P$; 
\(\mathbf{F}\) means $\lnot Q$ can be logically derived from $P$; \(\mathbf{U}\) means that, given $P$, the truth value of $Q$ cannot be determined.
\begin{tcolorbox}[fontupper=\linespread{0.7}\selectfont,]
{\footnotesize
$\blacktriangleright$ \textbf{Example:}\\
$<$Premises$>$ ($P$)\\
All cats are mammals. Tom is a mammal.\\
$<$Proposition$>$ ($Q$)\\
Tom is a cat.\\
$<$Answer$>$\\
\(\mathbf{Unknown}\).
}
\end{tcolorbox}

\subsection{Step 1: Translating}
The Translator maps natural language context to a structured symbolic representation. Specifically, given the raw premises $P=\{p_1,\dots,p_n\}$ and a proposition $Q$ to be decided, the Translator first produces the corresponding first-order logical forms $P'=\{p'_1,\dots,p'_n\}$ and $Q'$. Here, each $p'_i$ and $Q'$ is a well-formed formula over the standard first-order vocabulary (e.g., $\{\neg,\wedge,\vee,\rightarrow,\leftrightarrow,\forall,\exists,=\}\subseteq\mathcal{L}_{\mathrm{FO}}$). It then constructs a hybrid representation for downstream reasoning: for the premises, it creates the aligned set $P_c=\left\{\langle p_i, p'_i\rangle\right\}_{i=1}^{n}$; for the proposition, it forms $Q_c=\langle Q, Q'\rangle$. Finally, these components are combined into a structured symbolic representation $\mathcal{H}=\{P_c, Q_c\}$ for use in subsequent processing.

\begin{tcolorbox}[breakable, fontupper=\linespread{0.7}\selectfont]
{\footnotesize
$\blacktriangleright$ \textbf{Input:}\\
Please convert the context and question into formulas in First-Order Logic. \\
$<$Premises$>$ ($P$)\\ 
Fiona is quiet. \\
(... More premises ...) \\
$<$Proposition$>$ ($Q$)\\
Charlie is kind. \\[4pt]
$\blacktriangleright$ \textbf{Output:}\\[2pt]
\vspace{-0.3cm}
\begin{lstlisting}
<Structured symbolic representation> ($H$)
{
  "Premises ($P_c$)": [
    { 
      "statement": "Humans are mammals.", 
      "symbol": "$\forall$$x$ ($Human$ ($x$) $\rightarrow$ $Mammal$($x$))" 
    },
    (... More premises ..)
  ],
  "Proposition ($Q_c$)": { "statement": "There is an animal.",  "symbol": "$\exists$$x$ $A(x)$"}
}

\end{lstlisting}
}
\vspace{-0.1cm}
\end{tcolorbox}

\subsection{Step 2: Matrix-based Planner} 
From the structured representation $\mathcal{H}=\{P_c, Q_c\}$ given by the symbol translator, a matrix-based planner constructs an executable dependency plan $\Pi=(V,A)$. Here, $V=\{v_1,\dots,v_m\}$ is a finite set of reasoning steps. $A=[A_{ij}]_{i,j=1}^{m}\in\{0,1\}^{m\times m}$ is a dependency matrix. For each entry $A_{ij}$, if $A_{ij}=1$, then $v_i \prec v_j$, i.e., $v_i$ must be executed before $v_j$. The graph formed by $A$ is required to be a directed acyclic graph, which ensures the existence of a topological order to obtain a valid execution schedule and avoids execution difficulties caused by interleaved dependencies. Accordingly, we obtain a dependency plan $\Pi$ that is aligned with $P_c$ and $Q_c$ and has a clear structure, thereby balancing dependency and executability without sacrificing expressive power.

\begin{tcolorbox}[breakable, fontupper=\linespread{0.7}\selectfont]
{\footnotesize
$\blacktriangleright$ \textbf{Input:}\\
Please derive a step-by-step plan using First-Order Logic rules and a matrix-based method to determine the conclusion from the context.\\
<Structured symbolic representation> ($H$)\\
$\blacktriangleright$ \textbf{Output:}\\[2pt]
\vspace{-0.3cm}
\begin{lstlisting}
<Matrix-based Plan> ($\Pi$)
{
  "Plan ($V$)": {
    "1": { "content": "Define predicates." },
    "2": { "content": "Define constants: Monkeypox" },
    (... More plans ..)
    "11": { "content": "Make final judgment: True" }
  },
  "Matrix ($A$)":{
    [0,1,1,1,1,1,1,1,0,0,0],
    [0,0,1,1,1,1,1,1,0,0,0],
               $\cdots$
    [0,0,0,0,0,0,0,0,0,0,0]
    } 
}

\end{lstlisting}
}
\vspace{-0.1cm}
\end{tcolorbox}

\subsection{Step 3: Solver}

Given the symbolized premises $P_c=\left\{\langle p_i, p'_i\rangle\right\}_{i=1}^{n}$, the proposition $Q_c=\langle Q, Q'\rangle$, and the matrix-based plan \(\Pi=(V,A)\), the \textbf{Solver} executes the steps by following a linear order consistent with the dependency relations in the plan \(\Pi\). Concretely, the Solver refers to the premises \(P_c\) and performs inference according to the plan \(\Pi\), whose structure guarantees a valid topological execution order. After completing all steps, it uses the overall information of the reasoning process $\Gamma$.
Then it makes a three-valued decision on $Q_c$:
\begin{equation}
\hat y =
\begin{cases}
\mathbf{T}, & \text{if } \Gamma \vdash Q_c,\\
\mathbf{F}, & \text{if } \Gamma \vdash \neg Q_c,\\
\mathbf{U}, & \text{if } \Gamma \not\vdash Q_c \text{ and } \Gamma \not\vdash \neg Q_c.
\end{cases}
\end{equation}

The final output is the label \(\hat y\in\{\mathbf{T},\mathbf{F},\mathbf{U}\}\) together with execution log $\tau$ aligned with the dependency plan.

\begin{tcolorbox}[breakable, fontupper=\linespread{0.7}\selectfont]
{\footnotesize
$\blacktriangleright$ \textbf{Input:}\\
Please solve the question based on the context and a matrix-based plan.\\
<Structured symbolic representation> ($\mathcal{H}=\{P_c, Q_c\}$)\\
<Matrix-based Plan> ($\Pi$)\\
$\blacktriangleright$ \textbf{Output:}\\[2pt]
\vspace{-0.3cm}
\begin{lstlisting}
{
  "Execution log ($\tau$)": "Analyze steps according to the dependency matrix$\cdots$",
  "Final answer ($\hat y$)": {"T (True)"}
  }

\end{lstlisting}
}
\vspace{-0.1cm}
\end{tcolorbox}

\begin{table*}[!t]
\centering
\setlength{\tabcolsep}{4pt}
\renewcommand{\arraystretch}{1.1}
\fontsize{9pt}{9pt}\selectfont

\begin{adjustbox}{max width=\textwidth}
\begin{tabular}{@{}l|ccccc|ccccc@{}}
\toprule
\multirow{2}{*}{\textbf{Method}}
& \multicolumn{5}{c|}{\textbf{GPT-4o-mini}}
& \multicolumn{5}{c}{\textbf{GPT-4o}} \\
\cmidrule(lr){2-6}\cmidrule(lr){7-11}
& \textbf{PrOntoQA} & \textbf{ProofWriter} & \textbf{FOLIO} & \textbf{LogicalDeduction} & \textbf{AR-LSAT}
& \textbf{PrOntoQA} & \textbf{ProofWriter} & \textbf{FOLIO} & \textbf{LogicalDeduction} & \textbf{AR-LSAT} \\
\midrule
Std
& 56.00 & 35.00 & 65.20 & 57.00 & 22.94
& 77.20 & 53.33 & 64.71 & 71.64 & 29.00 \\
CoT
& \second{92.60} & 48.67 & \second{67.65} & 79.00 & 23.38
& \second{99.60} & 66.17 & 73.53 & 84.67 & 28.57 \\
Logic-LM
& 80.80 & 44.50 & 51.47 & \second{85.00} & 20.78
& 80.60 & \second{79.17} & 67.16 & \second{91.00} & 30.74 \\
SymbCoT
& 65.80 & \second{62.67} & 66.67 & 77.30 & \second{23.81}
& \second{99.60} & 76.50 & \second{74.02} & 85.67 & \second{38.53} \\
Aristotle
& 69.00 & 51.85 & -- & -- & --
& 75.95 & 74.12 & -- & -- & -- \\
Ours
& \best{98.00} & \best{71.50} & \best{71.57} & \best{86.33} & \best{37.23}
& \best{99.80} & \best{90.33} & \best{79.41} & \best{95.00} & \best{41.13} \\
\bottomrule
\end{tabular}
\end{adjustbox}

\caption{Accuracy of six methods across five reasoning datasets on GPT-4o-mini and GPT-4o. The second best score is \second{underlined} and the best is \best{bold}. “--” = N/A.}
\label{4o}
\end{table*}

\begin{table*}[!t]
\centering
\setlength{\tabcolsep}{4pt}
\renewcommand{\arraystretch}{1.1}
\fontsize{9pt}{9pt}\selectfont
\begin{adjustbox}{max width=\textwidth}
\begin{tabular}{@{}c|ccccc|ccccc|ccccc@{}}
\toprule

\multirow{2}{*}[-0.8em]{\textbf{Method}}
& \multicolumn{5}{c|}{\textbf{Qwen2.5-72B}}
& \multicolumn{5}{c|}{\textbf{DeepSeek-V3}}
& \multicolumn{5}{c}{\textbf{Kimi-K2}} \\
\cmidrule(lr){2-6}\cmidrule(lr){7-11}\cmidrule(lr){12-16}

& \makecell{\textbf{PrOnto}\\\textbf{-QA}}
& \makecell{\textbf{Proof}\\\textbf{-Writer}}
& \makecell{\textbf{FOLIO}}
& \makecell{\textbf{Logical}\\\textbf{-Deduction}}
& \makecell{\textbf{AR}\\\textbf{-LSAT}}
& \makecell{\textbf{PrOnto}\\\textbf{-QA}}
& \makecell{\textbf{Proof}\\\textbf{-Writer}}
& \makecell{\textbf{FOLIO}}
& \makecell{\textbf{Logical}\\\textbf{-Deduction}}
& \makecell{\textbf{AR}\\\textbf{-LSAT}}
& \makecell{\textbf{PrOnto}\\\textbf{-QA}}
& \makecell{\textbf{Proof}\\\textbf{-Writer}}
& \makecell{\textbf{FOLIO}}
& \makecell{\textbf{Logical}\\\textbf{-Deduction}}
& \makecell{\textbf{AR}\\\textbf{-LSAT}} \\

\midrule
Std
& 67.00 & 49.00 & 67.16 & 66.33 & \second{34.63}
& 72.20 & 58.00 & 63.73 & 67.00 & 29.44
& 53.20 & 48.67 & 65.69 & 70.00 & 36.36 \\
CoT
& \second{98.40} & 66.67 & \second{67.65} & \second{87.67} & 30.74
& \second{94.40} & 62.50 & 72.06 & 91.33 & 48.05
& 94.80 & 68.17 & 71.08 & 93.00 & 72.29 \\
Logic-LM
& 72.80 & 67.17 & 64.71 & 87.00 & 29.87
& 78.20 & 75.83 & 67.16 & 93.67 & 37.66
& \second{96.00} & 78.83 & 69.61 & \second{97.00} & 42.86 \\
SymbCoT
& 86.60 & 82.00 & 37.75 & 87.33 & 24.29
& 93.40 & \second{85.67} & \second{79.41} & \second{96.00} & \second{71.43}
& \best{99.40} & \second{84.50} & \second{79.90} & 87.33 & \second{72.73} \\
Aristotle
& 25.60 & \second{82.58} & -- & -- & --
& 73.60 & 59.42 & -- & -- & --
& 88.20 & 58.05 & -- & -- & -- \\
Ours
& \best{99.20} & \best{88.33} & \best{79.70} & \best{88.67} & \best{50.22}
& \best{99.80} & \best{92.50} & \best{81.77} & \best{99.67} & \best{74.89}
& \best{99.40} & \best{88.17} & \best{80.88} & \best{97.33} & \best{79.65} \\
\bottomrule
\end{tabular}
\end{adjustbox}

\caption{Accuracy of six methods across five reasoning datasets on three open models. The second best score is \second{underlined} and the best is \best{bold}. “--” = N/A.}
\label{cross}
\end{table*}

\subsection{Step 4: Feedback-driven replanner}

The Feedback-driven replanner treats the trace $\tau$ and the provisional outcome $\hat y$ as diagnostic evidence. The replanner inspects the dependency plan $\Pi=(V,A)$ and identifies structural inconsistencies. It then applies minimal dependency edits that preserve semantic validity and ensure topological executability. This procedure yields a revised plan $\Pi'=(V,A')$. A revision rationale accompanies the plan and documents the detected issues and their resolutions.

Re-execution under the revised plan $\Pi'$ yields a new reasoning trajectory and a definitive decision $y\in\{\mathbf{T},\mathbf{F},\mathbf{U}\}$. Thus, the entire feedback-driven process can be expressed concisely as:
\begin{equation}
\mathrm{FD\text{-}Replan}:\ (\mathcal{H},\Pi,\tau,\hat y)\ \mapsto\ (\Pi',\,\tau',\,y),
\end{equation}
where $\tau'$ is the updated execution log under $\Pi'$, and $y$ is the final three-valued answer consistent with the revised dependencies.

\begin{tcolorbox}[breakable, fontupper=\linespread{0.7}\selectfont]
{\footnotesize
$\blacktriangleright$ \textbf{Input:}\\
Please replan: 1) identify the deficiencies of the matrix-based plan and regenerate a revised plan; 2) subsequently re-execute the new plan.\\
<Structured symbolic representation> ($\mathcal{H}=\{P_c, Q_c\}$)\\
<Matrix-based Plan> ($\Pi$)\\
<Execution log> ($\tau$)\\
<Provisional answer> ($\hat y$)\\
$\blacktriangleright$ \textbf{Output:}\\[2pt]
\vspace{-0.3cm}
\begin{lstlisting}
{
  "Revised plan ($\Pi'$)":{
    "Corrected_plan": {
        "1": { "content": "Establish all initial facts from the premises." },
         (... More plans ..)
      },
      "Matrix": {
        [0,1,0,0],
          $\cdots$
        [0,0,0,0]
      }
}
  "Updated Execution log" ($\tau'$): "Establish Initial Facts from Premises$\cdots$"
  "Final answer ($y$)": "T (True)"
}

\end{lstlisting}
}
\vspace{-0.1cm}
\end{tcolorbox}

\section{Experiments}
\subsection{Experimental Setup}
\textbf{Dataset.} We evaluate on the same set of five logical–reasoning datasets: AR-LSAT~\cite{zhong2022analytical}, LogicalDeduction~\cite{srivastava2023beyond}, FOLIO~\cite{han2022folio}, Proof-\\Writer~\cite{tafjord2020proofwriter}, and PrOntoQA~\cite{saparov2022language}. These datasets span multiple symbolic representations and problem configurations, allowing us to test applicability and robustness across heterogeneous settings. PrOntoQA, ProofWriter, and FOLIO follow a first-order logic paradigm. LogicalDeduction and AR-LSAT are predominantly constraint-based. Together they cover different reasoning depths and choice cardinalities (e.g., T/F/U vs. multiple options). To avoid bias from differences in data splits, we adopt the same test subsets as Logic-LM and SymbCoT. During evaluation, any output with formatting noncompliance or missing fields is counted as incorrect. The detailed prompt templates are presented in Appendix~\ref{pt}.

\noindent{\textbf{LLMs.}} To comprehensively assess the effectiveness of the proposed method, we conduct comparative experiments on both closed- and open-source large language models. The closed-source models are GPT-4o (2024-11-20) and GPT-4o-mini (2024-07-18). The open-source models are Qwen2.5-72B, DeepSeek-V3, and Kimi-K2. To ensure a fair comparison and attribute performance differences primarily to the method rather than to prompt engineering, we evaluate each method on each model using the same prompt. Task instructions, exemplars, and output formatting are held fixed. Unless otherwise specified, all other inference settings are likewise kept constant.

\noindent{\textbf{Baselines.}} We adopt representative methods as baselines: Logic-LM~\cite{pan2023logic}, SymbCoT\cite{xu2024faithful}, and Aristotle~\cite{xu2024aristotle}. We also include two generic prompting styles—standard prompting and chain-of-thought (CoT) prompting that leverages the in-context learning capability of base LLMs. We conduct a unified evaluation on the five datasets described above. However, to ensure strictly fair comparability, the public implementation of Aristotle provides neither results nor reproducible configurations for FOLIO, AR-LSAT, or LogicalDeduction. Accordingly, we report their numbers only on PrOntoQA and ProofWriter; the remaining three entries are marked with “—”.

\subsection{Main Result}
\subsubsection{Overall Performance} 
As shown in Tables~\ref{4o} and ~\ref{cross}, MatrixCoT attains the best—or tied best—accuracy on both closed-source and open-source model families. Its variance across models and datasets is small. This indicates strong robustness and transferability. To ensure comparability, all methods use identical prompt templates and output constraints on each model. Under this setting, our method still leads. The gains therefore stem primarily from the method itself rather than from tailoring to a particular model or prompt.

\subsubsection{Method Differences and Empirical Observations} 
We observe that CoT achieves standout single-point results on certain models and datasets. We conjecture that this reflects extensive training exposure to chain-style narratives, which confers a format-alignment advantage on some tasks. However, this advantage fluctuates considerably across models and datasets. By contrast, although SymbCoT is also an LLM-driven framework, its textual planning and end-stage verification are sensitive to wording and sampling. It lacks machine-checkable cross-step dependencies and a process-level error-correction channel. As a result, it shows higher variance across models and is more prone to broken chains and omissions on long-chain or multi-branch cases. MatrixCoT employs a dependency-matrix-based plan that removes the uncertainty of “guessing dependencies” at execution time. It then absorbs issues exposed at runtime as structural repairs via minimal dependency edits driven by execution feedback. A unified structured output further reduces the risk of errors being counted due to format non-compliance. Consequently, it delivers a more stable and auditable performance profile across diverse base models.

\begin{figure}[h]
  \centering
  \includegraphics[width=\linewidth,keepaspectratio]{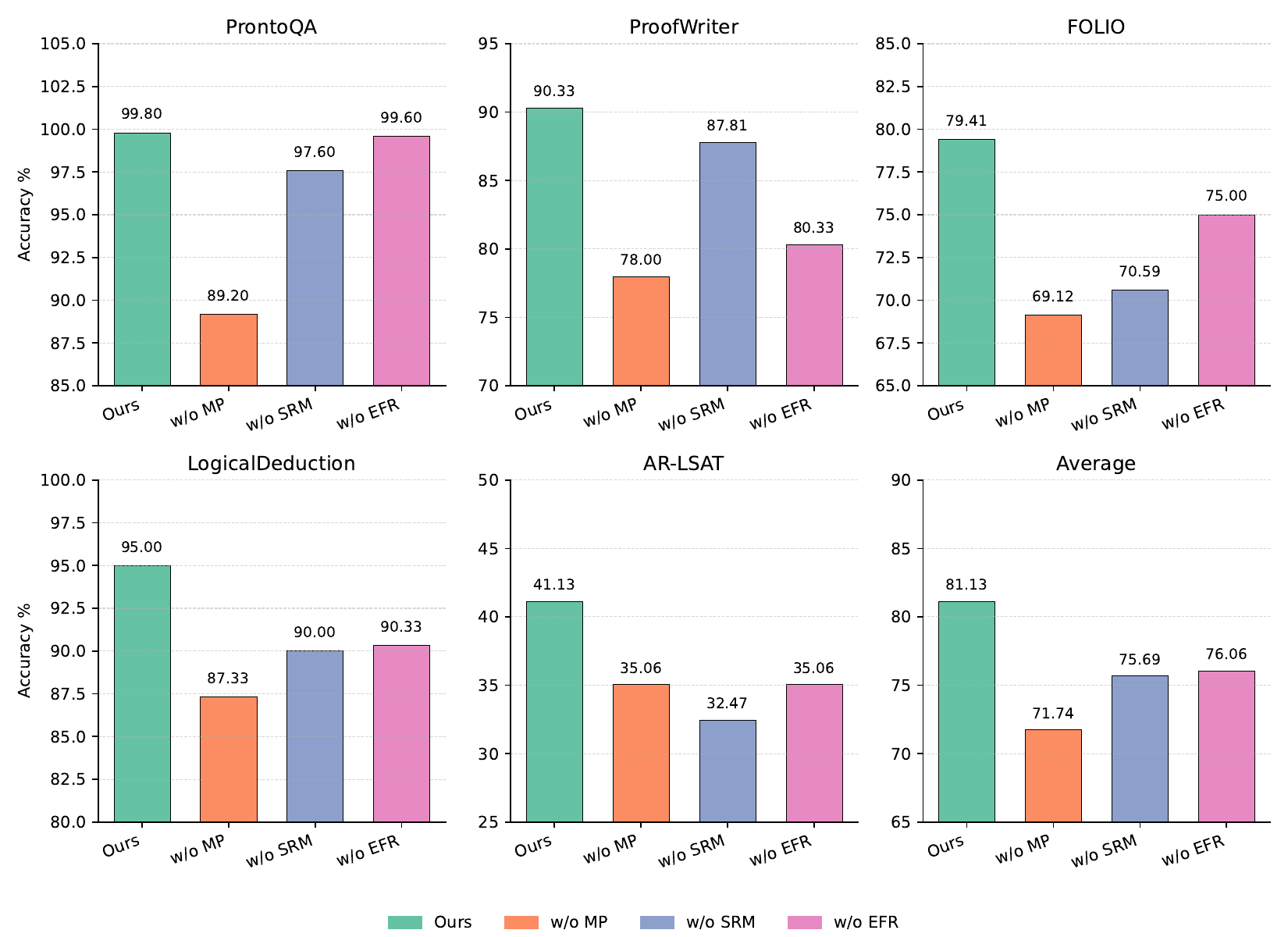}
  \caption{Ablation study. MP: Matrix-based plan; SRM: Structured data management; FDR: Feedback-Driven Replanner.}
  \label{ablation}
\end{figure}

\subsection{Ablation Study}
As shown in Table~\ref{ablation}, we conduct an ablation study of MatrixCoT. We remove three modules in turn: matrix-based planning (w\/o MP), structured reasoning data management (w\/o SRM), and execution-feedback replanning (w\/o EFR). The full configuration without any ablation consistently delivers the best performance. Removing MP lowers average accuracy by about 9 percentage points. Removing SRM or EFR each yields a drop of roughly 5–6 percentage points. The gain from EFR is more pronounced on long-chain or multi-branch tasks. On shallower reasoning tasks, sensitivity to EFR is lower, but MP and SRM still provide substantial benefits.

We hypothesize three main reasons for these improvements. \emph{1) SRM} converts free text into typed, indexable symbolic carriers. It explicitly constrains predicate signatures and referential links, curbing semantic drift and supplying verifiable anchors. \emph{2) MP} replaces a linear narrative with a precedence–successor dependency matrix. It turns “dependency discovery” into “dependency execution.” Through topological scheduling and transitive reduction, it preserves the minimal sufficient dependencies. This reduces broken chains, premature use, and omissions, and it facilitates cross-sample alignment and reuse. \emph{3) EFR} transforms execution traces and candidate answers into local minimal edits to the matrix and then re-executes. It feeds issues exposed at runtime back as provable repairs at the structural layer. Together, these three components elevate intermediate evidence from implicit narrative to computable, auditable artifacts, enabling an LLM-driven framework to achieve greater robustness, controllability, and consistency without external solvers.

\subsection{Analysis and Discussion}

\begin{figure}[t]
  \centering
  \includegraphics[width=0.95\linewidth,keepaspectratio]{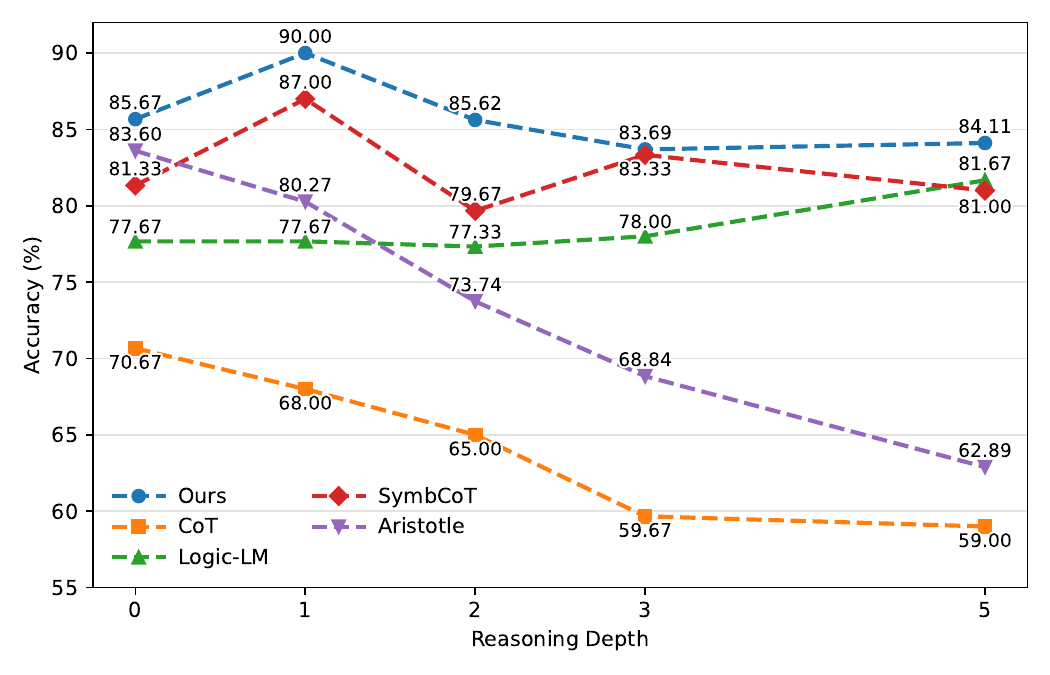}
  \caption{Accuracy on ProofWriter as reasoning depth increases.}
  \label{depth}
  \vspace{-0.4cm}
\end{figure}

\textbf{Reasoning Depth.} As presented in Fig.~\ref{depth}, we conduct a stratified evaluation of ProofWriter on DeepSeek-V3 across reasoning depths (from 0 to 5). MatrixCoT achieves the highest accuracy at every depth. Its accuracy declines only slightly as depth increases. Overall, it remains stable at roughly 84\%–90\%. By contrast, the baselines drop sharply with increasing depth. These results suggest that a dependency-based, matrix-structured plan provides explicit precedence constraints that guide execution and scheduling. It mitigates error accumulation in long-chain reasoning. Even at greater depths, it sustains robust performance.

\begin{figure}[t]
  \centering
  \includegraphics[width=1\linewidth,keepaspectratio]{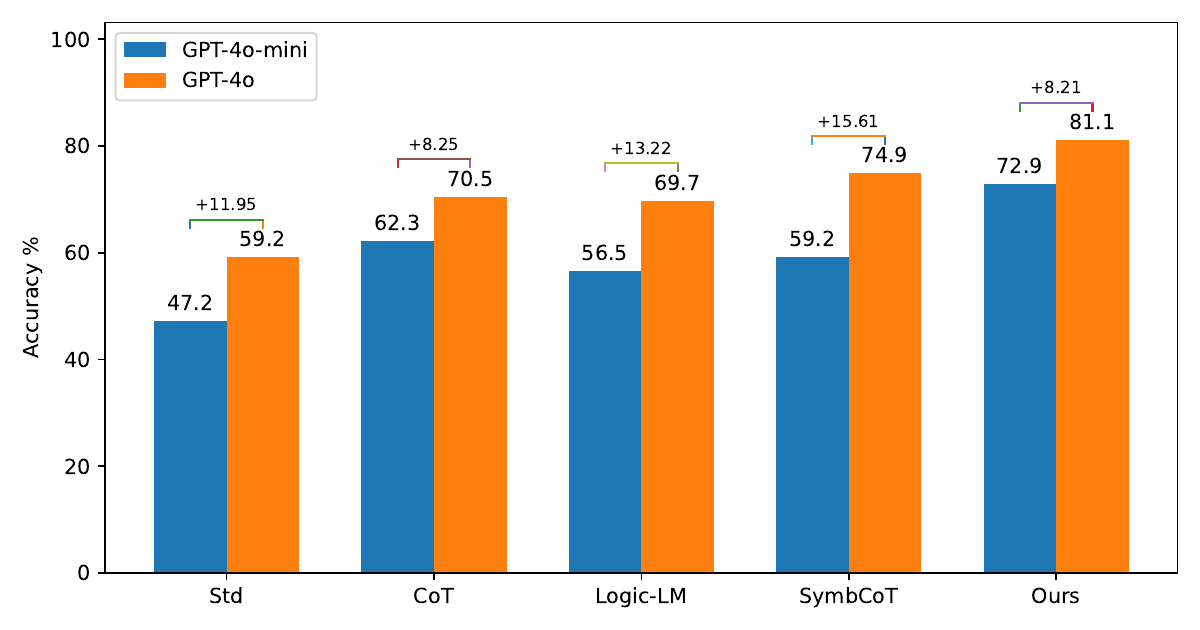}
  \caption{Improvement from GPT\textendash 4o-mini to GPT\textendash 4o.}
  \label{Figure 5}
    \vspace{-0.4cm}
\end{figure}

\textbf{Cross-Model Mean Performance.} As can be seen in Fig.~\ref{Figure 5}, on GPT-4o-mini, our mean accuracy is 72.93, ranking first. After upgrading to GPT-4o, it remains highest at 81.13. More importantly, the gain is only +8.21 percentage points—the smallest among methods covering all five datasets (Std +11.95; CoT +8.25; Logic-LM +13.22; SymbCoT +15.61). This indicates weaker reliance on increasing model size and smooth scaling on stronger models. The baselines achieve larger gains after the upgrade, yet their overall performance still does not exceed ours. This further shows greater stability and robustness across model capacities. Note that Aristotle covers only two datasets; its mean and gain are not directly comparable and are therefore excluded from the statistics.

\textbf{Error Analysis.} Reliable symbolic reasoning depends on three capabilities: 1) Information acquisition and structuring with target-relevant content; 2) Path selection and plan orchestration by explicitly modeling precedence–successor constraints and maintaining global consistency; and 3) Executing tasks with constraints and accuracy by performing semantically correct derivations according to the specified plan. MatrixCoT strengthens the first two via structured reasoning data management and a matrix-based plan. Structured reasoning data management provides a typed, alignable symbolic base. The matrix plan fixes executable precedence relations through a dependency matrix $A$. Together, they make “finding the right information and laying out the correct path” a high-probability event. On this basis, execution is typically smoother and yields robust gains. For models with weak instruction-following ability, execution can still fail. It occurs even when the plan is logically complete and its dependencies are explicit. The failures often stem from an inadequate grasp of the dependency constraints. Thus, MatrixCoT reduces reliance on free-text planning and ad hoc search, but it does not substitute for the base model's core execution ability. As a model's basic comprehension and compliance improve with scale, it exploits the matrix plan more effectively, and the gains become more stable.

\section{Conclusion}
In this paper, we introduce MatrixCoT, a structured CoT framework that integrates a matrix-based method and a feedback-driven replanning mechanism. Specifically, we enforce a unified structured output format and normalize solution steps as typed symbolic representations, while explicitly encoding cross-step constraints via a dependency matrix. We also derive a verifiable, executable stepwise plan, and introduce a feedback-driven replanning mechanism, which can repair broken chains and correct deviations. The entire procedure operates without any external solver, yet improves robustness, interpretability, and controllability. Extensive experiments are conducted on five standard benchmarks and five LLMs (both closed- and open-source), demonstrating stable and competitive performance across all settings.

\begin{acks}

This work was supported in part by the Joint Funds of the National Natural Science Foundation of China under Grant U25A20436; by the Guangxi Key Research \& Development Program (FN2504240036, 2025FN96441087); by the National Natural Science Foundation of China (NSFC) (62372047, 62302049); by the Natural Science Foundation of Guangdong Province (2024A1515011323); by the Supplemental Funds for Major Scientific Research Projects of Beijing Normal University, Zhuhai (ZHPT2023002); by the Fundamental Research Funds for the Central Universities; by the Industry–University Cooperation Collaborative Education Project of the Ministry of Education (240904497110437); and by the Higher Education Research Topics of the Guangdong Association of Higher Education in the 14th Five-Year Plan (24GYB207). We acknowledge the support of the Interdisciplinary Intelligence Super Computer Center of Beijing Normal University at Zhuhai.

\end{acks}

\bibliographystyle{ACM-Reference-Format}
\bibliography{sample-base}

%


\appendix
\section{Appendix}
\subsection{Prompt Templates}~\label{pt}
We present a brief description of the prompt setup used on the ProofWriter dataset. A total of six prompts are employed, organized into four modules:  Translator,  Matrix-based Planner,  Solver, and Replanner (which includes Replanning and Re-Execution). These prompts together support the complete process from natural language propositions to automated reasoning and verification.

\begin{figure*}[t]
    \centering
    \includegraphics[width=\linewidth]{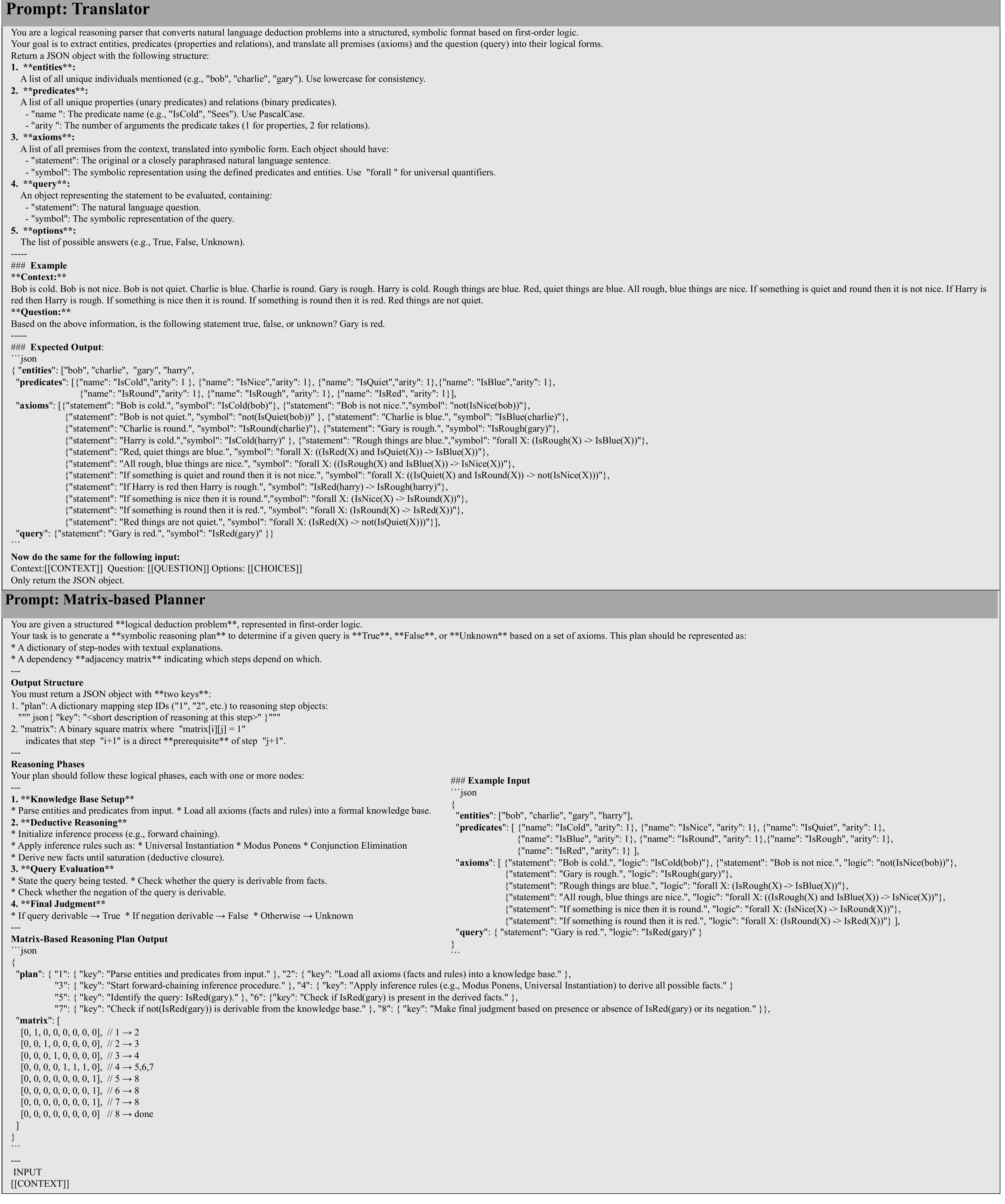}
    \label{p1}
\end{figure*}

\begin{figure*}[h]
    \centering
    \includegraphics[width=\linewidth]{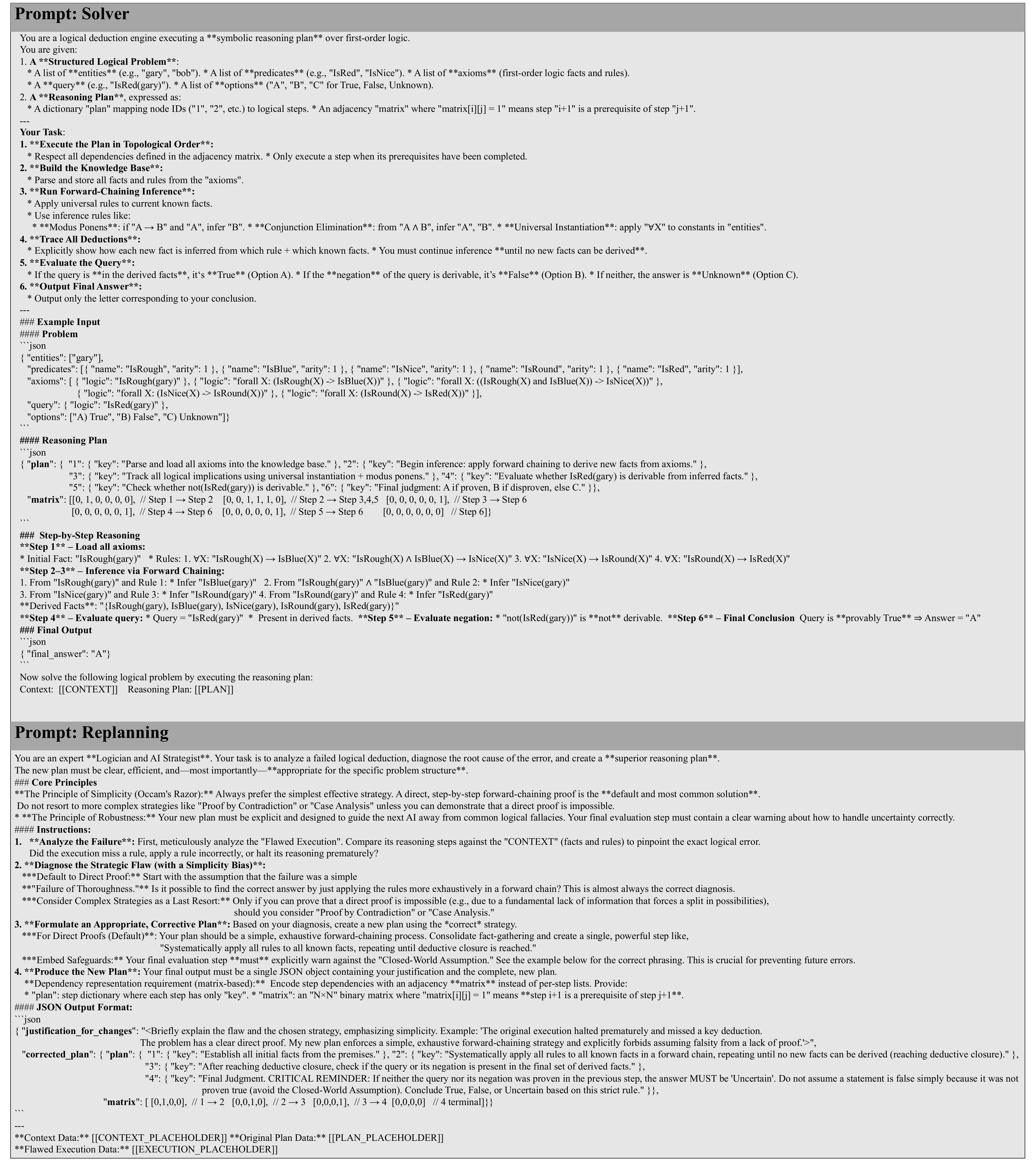}
    \label{p2}
\end{figure*}

\begin{figure*}[h]
    \centering
    \includegraphics[width=\linewidth]{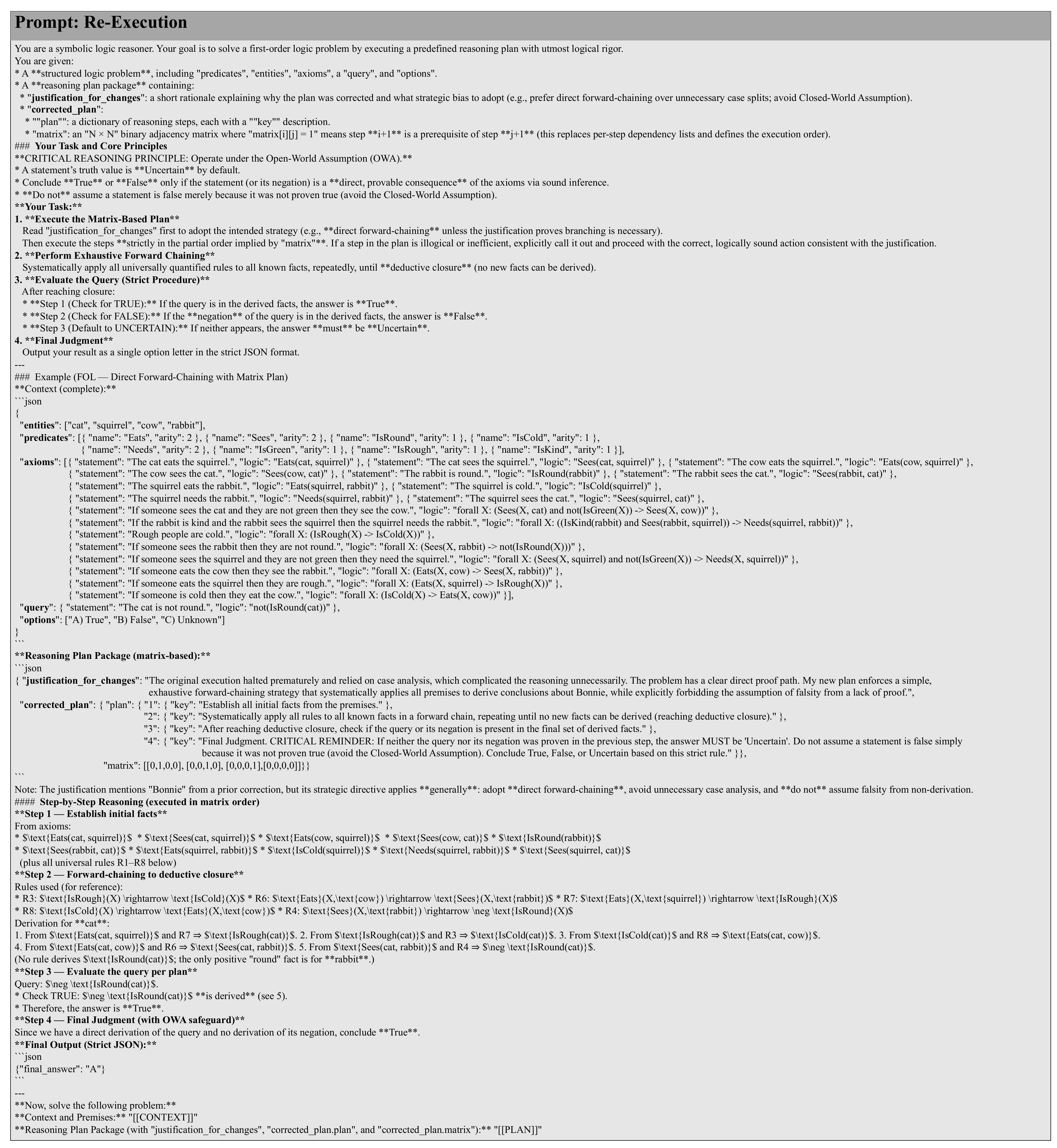}
    \label{p3}
\end{figure*}

\end{document}